\documentclass[]{LTJournalArticle}

\runninghead{Hybrid Feature- and Similarity-Based Models} 

\footertext{ } 

\title{Hybrid Feature- and Similarity-Based Models for Joint Prediction and Interpretation} 

\author{%
	Jacqueline K Kueper\textsuperscript{1,2}\thanks{Corresponding author: \href{mailto:jkueper@uwo.ca}{jkueper@uwo.ca}}, Jennifer Rayner\textsuperscript{3} and Daniel J Lizotte\textsuperscript{1,2}
}

\date{\footnotesize\textsuperscript{\textbf{1}}Department of Epidemiology and Biostatistics, The University of Western Ontario\\ \textsuperscript{\textbf{2}}Department of Computer Science, The University of Western Ontario\\ \textsuperscript{\textbf{3}}The Alliance for Healthier Communities}

\renewcommand{\maketitlehookd}{%
	\begin{abstract}
		\noindent Electronic health records (EHRs) include simple features like patient age together with more complex data like care history that are informative but not easily represented as individual features. To better harness such data, we developed an interpretable hybrid feature- and similarity-based model for supervised learning that combines feature and kernel learning for prediction and for investigation of causal relationships. We fit our hybrid models by convex optimization with a sparsity-inducing penalty on the kernel. Depending on the desired model interpretation, the feature and kernel coefficients can be learned sequentially or simultaneously. The hybrid models showed comparable or better predictive performance than solely feature- or similarity-based approaches in a simulation study and in a case study to predict two-year risk of loneliness or social isolation with EHR data from a complex primary health care population. Using the case study we also present new kernels for high-dimensional indicator-coded EHR data that are based on deviations from population-level expectations, and we identify considerations for causal interpretations.  
	\end{abstract}
}

\begin{document}

\maketitle 


\section{Introduction}
\label{sec:intro}

Data from health care settings often include known, informative features as well as additional data that may be useful for the task but are challenging to summarize into meaningful features due to size or complexity. For example, electronic health records (EHRs) capture client characteristics (e.g., year of birth) in structured fields and record information arising from each encounter for that client (e.g., time-stamped diagnosis and procedure codes) in dynamic tables. The former may be well suited for features while the latter high-dimensional, variable-length data may be better represented in terms of similarity to other people in the database. 

Feature- and similarity-based learning approaches have complimentary characteristics. Feature-based approaches, such as logistic regression (\emph{LR}), tend to be more familiar to end-users, less susceptible to overfitting, and easier to interpret; however, not all valuable information can be captured with features and model performance may suffer from underfitting, especially for heterogeneous populations. In contrast, similarity-based approaches such as multiple kernel learning have a higher computational cost and cannot be interpreted on their own for the purpose of causal inference, but they can incorporate more complex or time-varying data that may account for additional variability in the outcome, or can be useful when good feature representations are unknown \citep{shawe-taylor2004, conroy2017, gonen2011}.

Our \textbf{primary objective} was to develop an intrinsically interpretable \emph{hybrid feature- and similarity-based model} (\emph{HFSM}). Our novel model form supports interpretations of feature coefficients while reaping additional benefits from similarity based approaches, such as improved absolute risk prediction while maintaining traditional feature interpretations, or adjustment for complex confounders in clinical epidemiology modelling. We evaluated two variations of \emph{HFSM} with 1) synthetic data and 2) EHR data from a complex primary health care population, finding our approach can outperform solely feature- or similarity-based methods while retaining or enhancing interpretability. Our \textbf{contributions} include: 
\begin{itemize}[noitemsep, topsep=0pt]
    \item A novel hybrid feature- and similarity-based model based on feature and kernel learning that supports joint prediction and interpretation tasks. 
    \item A new framework for developing kernel functions in terms of the presence of rare and absence of common characteristics. 
    \item A case study on predicting social isolation and loneliness for a complex primary health care population. \emph{HFSM} predictive performance was better or comparable to simple models and opaque complex models, and may be interpreted to learn about social isolation and loneliness related characteristics as captured in primary health care. 
\end{itemize}


\section{Related Work}
\label{sec:relatedWork}

\paragraph{Integrating multiple types of data.} 
\citet{fan2017} developed \emph{RIT-UA} for recommender systems, first computing similarity scores for 1) user attribute data and 2) time-varying, high-dimensional data arising from user interactions with a system, e.g., histories of movie viewings or ratings. Predictions are made through a weighted linear combination of the two scalar scores. \citet{lian2015} proposed a multiview learning framework whereby feature and/or similarity matrices are each assumed to contribute a different ``view" of a dataset, and a shared latent factor matrix is learned as a global representation. Multiple kernel learning \citep{shawe-taylor2004, conroy2017, gonen2011}, which can be incorporated into \emph{HFSM}, is a special case. Like \emph{HFSM}, the above approaches use different techniques on different data types; however, while they both combine the resulting information into a generic form, \emph{HFSM} maintains a separate feature matrix to facilitate better interpretability.

\paragraph{Combining model types.} \citet{hothorn2010} developed \emph{mboost}, which combines penalized least square estimates and/or regression tree base learners in a weighted additive model structure. Each component is fit on all or a subset of data and can be interpreted separately \citep{hothorn2021, hothorn2010}. Our sequentially-optimized \emph{HFSM} approach is similar, but uses different components and does not employ an overall weight for each model component. \citet{sigrist2021} developed \emph{KTBoost}, which learns both a regression tree and a reproducing kernel Hilbert space regression function on all available data in each iteration, adding whichever is expected to result in better performance to the ensemble of base learners. \emph{KTBoost} does not segregate data or allow feature and kernel coefficients to be jointly optimized as in our simultaneous \emph{HFSM} approach. The gradient boosting technique \emph{XGBoost} fits new decision tree models to account for residual errors from previous models until performance stops improving \citep{chen2016}. \emph{XGBoost} has demonstrated excellent predictive performance in several settings, but as with the other boosting techniques, the focus is on predictions. Our \emph{HFSM} is parametric with a fully convex objective function, which supports reproducibility and clinical interpretation of the model.

\paragraph{Kernel functions for clinical data.} \citet{shawe-taylor2004} review standard kernels for sets or strings (e.g., intersection kernel, agreement kernel) that could be applied to indicator EHR data, but are solely based on equally-weighted present variables. \citet{klenk2010} used regression techniques to weight the importance of input variables for assessing overall similarity between two clients. \citet{belanche2013} used probability of occurrence in the training data to weight input variables such that rare ones have more influence. We also calculate similarity non-uniformly across input variables by applying weights derived from training data; however, we additionally explore similarity due to shared absence of common variables. Our kernel approach is also related to work that applies different functions to different types of input variables, e.g., ordinal versus nominal \citep{daemen2012, chan2010}, and to work on learning composite kernels as a structure discovery problem \citep{duvenaud2013}. 


\section{Hybrid Feature and Similarity-based Model}\label{sec:methods}

We address prediction tasks where the outcome ${y}(o)$ for an observation $o$ is explained partly by features $\vec{\phi}(o)$ (i.e.\ a fixed-dimension vector of properties of $o$) and partly by more complex information $\Psi(o)$ (e.g., high-dimensional, time-varying, variable-length data). When developing a model, similarity-based approaches may be advantageous over feature-based approaches when there are a large number of data elements or characteristics to consider and/or when the proper way to enter data into a model is unknown. For example, in a primary health care setting each observation could be a client and the outcome of interest a condition or situation, such as diabetes or food insecurity, that the client is at risk for and early intervention may help to prevent. In this setting, $\phi(o)$-type information may include sociodemographic characteristics and core diagnoses known to be associated with the outcome while $\Psi(o)$-type information may include years of encounter data representing the subset of thousands of possible tests, diagnoses, and procedures that the client has received in their lifetime. Additional technical background on feature- and similarity-based approaches is provided in Appendix \ref{appendix:AO3-bkgd}. 

Our proposed \emph{HFSM} combines feature-based and similarity-based components with an additive model structure. Prediction $\hat{y}(o)$ for observation $o$ is given by
\begin{align*}
\hat{y}(o) &= h\bigl(\sum_{j}\phi_{j}(o)\beta_{j}+\\
    & \sum_{i}\alpha_{i}\sum_{l}k_{l}\bigl(\Psi_{l}(o),\Psi_{l}(o_{i})\bigr)\bigr)\\
&= h\bigl(\boldsymbol{\phi}(o)^{T}\boldsymbol{\beta}+\boldsymbol{k}(o)^{T}\boldsymbol{\alpha}\bigr)
\end{align*}
\noindent where $j$ are features; $i$ are training data observations; $l$ are kernel domains, if multiple; and $h$ is a monotonic function, e.g., sigmoid or identity. Herein we use the sigmoid function such that the estimated probability of an outcome is based on 1) $o$'s feature values and the corresponding coefficients ($\beta_j$) and 2) $o$'s similarity to training data observations and each of their overall influence ($\alpha_{i}$). Similarity to clients with positive $\alpha$ will increase the predicted probability while similarity to clients with negative $\alpha$ will decrease it. 


\subsection*{Fitting and interpretation.} 
To train the model, we optimize a penalized log likelihood training criterion given by
\begin{align*}
\textit{LL}(\boldsymbol{\beta},\boldsymbol{\alpha};\lambda) = \bigl( \sum_i y_i (\boldsymbol{\phi}(o_i)^{\mathsf{T}}\boldsymbol{\beta} + \boldsymbol{k}(o_i)^{\mathsf{T}}\boldsymbol{\alpha}\bigr) \\ 
 - \log\bigl(1 + e^{\boldsymbol{\phi}(o_i)^{\mathsf{T}}\boldsymbol{\beta} + \boldsymbol{k}(o_i)^{\mathsf{T}}\boldsymbol{\alpha}}) \bigr)/n - \lambda ||\boldsymbol{\alpha}||_1
\end{align*}

The L1-penalty on $\vec{\alpha}$ controls overfitting and produces a sparse model whose kernel component only depends on a subset of the training data; this is different from the original kernel logistic regression formulation which penalizes the norm of the regression function in its Hilbert space but does not induce sparsity \citep{zhu2005}. Training $o_i$ that maintain non-zero $\alpha$ can be thought of as ``representatives" for groups of similar clients. We solve this problem using the convex programming language \texttt{cvxpy} in Python \citep{diamond2016, agrawal2018}. An illustrative example relating the hybrid model to a special case of kernel logistic regression and Python code are provided in Appendix \ref{apd:AO3-PDexample} and \ref{apd:AO3-modelcode}, respectively. Required memory for model fitting, assuming $n$ clients and $m$ features, is $\mathcal{O}[mn + n^2]$ for \emph{HFSM} as compared to $\mathcal{O}[mn]$ for \emph{LR} and $\mathcal{O}[n^2]$ for kernel logistic regression (\emph{KLR}).

We consider two variations on fitting \emph{HFSM} that have different interpretations: 1) \emph{HFSM-Sequential} (\emph{HFSM-Seq}), which learns the feature coefficients fixing $\vec{\alpha} = 0$ and then fixes the learned feature coefficients while learning the kernel coefficients, and 2) \emph{HFSM-Simultaneous} (\emph{HFSM-Sim}), which learns the feature and kernel coefficients jointly. \emph{HFSM-Sim} is expected to result in better predictive performance since there is more flexibility to maximize the objective function, but the resulting model has a more complex feature coefficient interpretation. 

\paragraph{Feature coefficient interpretation.} In \emph{HFSM-Seq}, the feature coefficients represent their impact on the outcome adjusted for all of the other features in the model but \textit{averaged} over the information in the kernel, whereas for \emph{HFSM-Sim} feature coefficients are additionally \textit{adjusted} for the information in the kernel. If the feature and kernel matrices are orthogonal, the models produced by the two procedures will be identical. A series of illustrative examples in Appendix \ref{apd:interpEx} contrast the performance and interpretation of \emph{HFSM-Seq} and \emph{HFSM-Sim}. Across four scenarios that varied the causal role of a kernel with respect to two features, \emph{HFSM-Sim} always had the best predictive performance, but feature coefficient estimates could be closer to the truth, further from the truth, or similar to the feature coefficients learned by \emph{HFSM-Seq}. In practice the direction of bias, if any, may be hard to determine. This uncertainty is analogous to situations with solely feature-based approaches where the relationships between the features and the outcome are unknown, or when automatic feature selection methods are used \citep{brookhart2010, greenland2001, shrier2008, austin2004}. However, when there is adequate understanding about the information in the kernel, \emph{HFSM-Sim} can provide an interpretation advantage by adjusting for more complex potential confounding than can be captured through features. When information captured by the kernel is uncertain, \emph{HFSM-Seq} can be used to maintain straightforward feature coefficient interpretation while improving absolute risk prediction through addition of the kernel. 


\paragraph{Kernel coefficient interpretation.} The kernel coefficients $\vec{\alpha}$ may also be informative in and of themselves. Whereas applying an L1-penalty to features is a form of feature selection, applying an L1-penalty to $\vec{\alpha}$ selects ``representative observations" to include while adjusting for the features. The higher the penalty, the fewer observations are allowed. The most influential clients in the training data (highest magnitude $\alpha$) can be investigated to explore kernel behaviour. 


\section{Simulation Study}\label{sec:SS} 

We performed an ablation study across three scenarios that varied the relative importance of feature- and kernel-based data. We followed the ADEMP framework for planning and reporting \citep{morris2019}.


The data generating mechanism was a parametric model with four binary features and additional complex information $M$ that cannot be well represented in a linear model: $P(Y) = \varsigma(\beta_{0} + 0.3X_{1} + 0.4X_{2} + 0.6X_{3} + 0.7X_{4} + \delta M)$. $M$ is defined as the {Monk-1} problem, a non-linear function of six additional variables \citep{thrun1991}. 
 

Coefficients were set so if $\beta_{0}=0$, $\delta = 0$, and $\sum_{m=1}^{4} \beta_{m} \tilde = 2$ then $P(Y)$ ranges from 0 to 0.88. We manipulated $\delta$ and then set $\beta_{0}$ to bring the outcome prevalence below 50\% to be more similar to most clinical outcomes.
Three scenarios were set up with 10,000 observations generated from each, which is similar to the number of clients expected across a few small primary health care clinics: 
\begin{enumerate}[noitemsep, topsep=0pt]
  \item Kernel had a similar effect to a single feature: $\delta = \mathrm{mean}(\vec{\beta})$ and $\beta_{0} = -1.5$
  \item Kernel had a similar effect to the set of features: $\delta = \mathrm{sum}(\vec{\beta})$ and $\beta_{0} = -2.1$
  \item Kernel had a larger effect than the set of features: $\delta = 2\cdot \mathrm{sum}(\vec{\beta})$ and $\beta_{0} = -3.2$
\end{enumerate}
To compare models, we used 5-fold nested CV with outer fold training data split 75/25. Folds were consistent across models, and random seeds were re-set between scenarios. Any hyperparameters were selected by grid search of AUROC using inner validation data. Models were re-trained with selected hyperparameters on all outer fold training data and evaluated on outer test data. Our primary metric was AUROC; secondary metrics included AUPRC, calibration plot slopes and intercepts, and outer fold training time. We ran folds in parallel using the python package \texttt{multiprocessing} \citep{mckerns2010, mckerns2011}.

We compared four models: 1.\ Feature only: \emph{LR}, 2.\ Similarity only: \emph{KLR}, 3.\ \emph{HFSM-Seq}, 4.\ \emph{HFSM-Sim}, and 5. a ``best possible model" with known coefficients applied to all data. For similarity-containing models, we applied the RBF kernel to the six categorical variables of the Monk-1 data, as done in previous work \citep{belanche2013, marquez2014}. Candidate hyperparameter values for $\sigma$ (0.01, 0.1, 1) provided a range of similarity patterns (Appendix \ref{apd:AO3-rbfSigSelect}), and five values for $\lambda$, ranging from 0.001 to 1, gave a range of possible regularization.

\subsection*{Results for simulation study} 

As expected, hybrid models performed similar to or better than single component models across all scenarios, with the most advantage on discrimination and precision-recall performance for the second scenario (Table \ref{tab:table-ss-keyResults}). These findings show the advantage of using \emph{HFSM} when both feature- and kernel-based data are important and when there is uncertainty about their relative importance.  
Selected hyperparameters and learned parameters are in Appendix \ref{apd:AO3-SSextraRes}. 

\begin{table*}[!htb] 
    \centering
    \caption{Simulation study results.} 
    \label{tab:table-ss-keyResults}
    \begin{threeparttable}
\begin{tabular}{lccccc}
\toprule
  & LR & KLR & HFSM-Seq & HFSM-Sim & Best\\
\midrule
\addlinespace[0.3em]
\multicolumn{6}{l}{\textbf{Scenario 1: Kernel data had similar effect to a single feature}}\\
\hspace{1em}AUROC & 0.647 & 0.504 & 0.648 & 0.647 & 0.655\\
\hspace{1em}AUPRC & 0.571 & 0.436 & 0.572 & 0.573 & 0.581\\
\hspace{1em}Calibration Slope & -0.025 & -0.362 & -0.032 & -0.013 & -0.008\\
\hspace{1em}Calibration Intercept & 1.035 & -0.517 & 1.038 & 1.031 & 0.989\\
\hspace{1em}Time (hours) & $<$ 1 & 6.553 & 7.815 & 7.177 & \\
\addlinespace[0.3em]
\multicolumn{6}{l}{\textbf{Scenario 2: Kernel data had similar effect to the set of features}}\\
\hspace{1em}AUROC & 0.614 & 0.712 & 0.725 & 0.726 & 0.781\\
\hspace{1em}AUPRC & 0.587 & 0.672 & 0.708 & 0.710 & 0.759\\
\hspace{1em}Calibration Slope & -0.001 & 0.043 & 0.044 & 0.008 & 0.027\\
\hspace{1em}Calibration Intercept & 0.993 & 1.415 & 1.305 & 1.257 & 1.017\\
\hspace{1em}Time (hours) & 0.001 & 9.901 & 10.574 & 9.316 & \\
\addlinespace[0.3em]
\multicolumn{6}{l}{\textbf{Scenario 3: Kernel data had a larger effect than the set of features}}\\
\hspace{1em}AUROC & 0.558 & 0.872 & 0.877 & 0.877 & 0.903\\
\hspace{1em}AUPRC & 0.538 & 0.825 & 0.840 & 0.846 & 0.879\\
\hspace{1em}Calibration Slope & -0.001 & 0.010 & 0.012 & -0.058 & 0.018\\
\hspace{1em}Calibration Intercept & 0.980 & 1.557 & 1.575 & 1.534 & 1.018\\
\hspace{1em}Time (hours) & $<$ 1 & 7.148 & 7.419 & 5.111 & \\
\bottomrule
\end{tabular} \footnotesize{Results are averaged across five outer folds. \emph{Legend:} AUPRC = Area Under Precision Recall Curve, AUROC = Area Under Receiver Operating Characteristic Curve, Best = Hardcoded true coefficients applied to all data, HFSM-Seq = Hybrid Model - Sequential Fit, HFSM-Sim = Hybrid Model - Simultaneous Fit, KLR = Kernel logistic regression, LR = Logistic Regression.}
     \end{threeparttable}
\end{table*}

\section{Clinical Case Study} 

The Alliance for Healthier Communities provides inter-professional, team-based primary health care at Community Health Centres (CHCs) across Ontario, Canada \citep{albrecht1998, allianceforhealthiercommunities2020}. All CHCs record sociodemographic information (e.g., birth date, education) and appointment details (e.g., care provider type, diagnosis codes) in a centralized EHR database. We used data from 2009-2019 to predict two-year risk of first incidence loneliness or social isolation for middle-aged ongoing primary care clients being served by the ``urban-at-risk" (UAR) peer group of CHCs. This subgroup of CHCs provides care to clients with pre-existing substance use, homelessness, or mental health challenges. Social isolation and loneliness are increasingly recognized as serious health challenges, with research focussing on sequelae and comorbidities in older adults, finding associations with several poor health outcomes \citep{altschul2021, holt-lunstad2017, koning2017, nationalacademiesofsciencesengineeringandmedicine2020, nevarez-flores2021, nicholson2012, worldhealthorganization2021, lim2020}. CHCs provide a range of services that may help mitigate risk, such as social prescribing initiatives \citep{nowak2021, mulligan2020}. Our study was approved by Western University ethics board (project ID 111353). 

\paragraph{Cohort} Appendix \ref{apd:AO3-Cohort} has extended cohort details. For each included client, we randomly selected a two-year prediction interval across the client care history, excluding the period within one year of the first recorded event because the first year of care has a distinct risk profile \citep{kueper2022a}. Intervals were labeled positive if they contained a first incidence of loneliness or social isolation. Baseline predictors were from the first recorded event to the interval start date. 



\paragraph{Features} We identified 19 features based on literature \citep{altschul2021, doryab2019, koning2017, nicholson2012, nationalacademiesofsciencesengineeringandmedicine2020, holt-lunstad2017, worldhealthorganization2021}, input from Alliance stakeholders, and data availability. For features with less than 1\% missingness we applied complete case analysis, otherwise we used a missingess indicator. Features from appointment-associated diagnosis codes \citep{worldhealthorganization2020, bernstein2009} with under 1\% prevalence were excluded, but we performed indirect standardization to assess sex-adjusted risk in associated subpopulations. 
We also constructed a feature for the number of baseline chronic conditions \citep{fortin2017}, scaled to 0-1 range, to capture general clinical complexity. 

\paragraph{Kernel inputs} In addition to the identified features, general clinical complexity may be positively associated with the outcome but is challenging to meaningfully capture as features. Therefore, we used three types of appointment-associated data as candidate \emph{kernel inputs}: 1) provider type(s) involved in care (e.g., nurse practitioner, social worker), 2) service type(s) provided during an appointment (e.g., assessment, treatment management), and 3) both 1 and 2. There are many ways data could be pre-processed and combined for kernel inputs; we worked with sets of codes recorded during baseline care.

\paragraph{Common absence/rare presence kernels}

For the case study, we developed kernel functions with the following properties, based on domain knowledge:
\begin{enumerate}[noitemsep, topsep=0pt]
  \item Two clients who both have or do not have a specific code should be more similar than when only one has it. 
  \item Two clients who do not have a code that is common in the population of interest should be more similar than two people who both have the code present. 
  \item Two clients with a rare code present should be more similar than otherwise, but sharing in the absence of rare codes should not have large similarity impacts.  
\end{enumerate}
We developed kernel functions based on \citet{gower1971}'s work on the coefficient of similarity. The similarity $S_{i,j,c}$ between two clients $i$ and $j$ based on code $c$ ranges from 0 (no similarity) to 1 (perfect match): $S_{i,j} = {\sum_{c=1}^v S_{i,j,c}w_{c}} / {\sum_{c=1}^v \delta_{i,j,c}w_{c} }$. The 0-1 indicator $\delta_{i,j,c}$ represents whether a comparison is possible (code is present for one or both clients), and $w_{c}$ are code weights. 


The first property is achieved by the \emph{Jaccard (J)} similarity, obtained by setting setting $w_{c}=1$ for all $c$; when both sets are empty we set $S_{i,j}$ to 1. The second property can be achieved by reverse coding common data elements and assigning positive weights only to common codes. The third property can be achieved using presence-based coding and assigning positive weights only to rare codes. We used a cut-off based on prevalence in the training data to define common (prevalence $\ge$ 0.70) and rare (prevalence $<$ 0.30) codes; those above/below the threshold are assigned $w_{c}=1$ and remaining are assigned $w_{c}=0$. The resulting kernel, \emph{Jaccard Common Rare extension (J-CR)} adds together the ``common absence" and ``rare presence" similarity scores.

\paragraph{Application 1: Prediction} 
To assess predictive performance, we used a similar nested CV procedure as for the simulation study, with 80/20 splits to define inner training/validation data for each of five outer folds, and an AUROC-based grid search for the best hyperparameters on the inner loop. We compared the same four core models (LR, KLR, HFSM-Seq, HFSM-Sim) alongside \emph{LR-E} (\emph{LR} with the extra chronic condition count feature) and a more complex model (\emph{XGBoost}), which included the features and all kernel data as dummy variables. Hyperparameters included L1 penalty strength (0.0001, 0.001, or 0.01), kernel data inputs (providers involved, service types, or both), and kernel function (\emph{J} or \emph{J-CR}).

\paragraph{Application 2: Interpretation} 
To demonstrate model interpretability, we re-trained \emph{HFSM-Seq} and \emph{HFSM-Sim} on the entire eligible cohort using \emph{J} kernel function on both types of data with the mode of the selected L1 penalty from Application 1, divided by five to scale for the increase in amount of data. We compared feature coefficients between the two models. To examine the type of information captured by the kernel after accounting for features, we used \emph{HFSM-Seq} to split the cohort by positive, negative, and zero-valued $\alpha$. Feature-based characteristics were compared across the three strata using table-based summaries. Kernel-based characteristics were explored using non-negative matrix factorization (NMF) with five topics, using Python package \texttt{sklearn.decomposition.NMF} with Kullback-Leibler divergence, to each of the three strata \citep{pedregosa2011}. 


\subsection*{Results for clinical case study} 

There were 5,070 eligible clients, of whom 5.4\% (n=276) had the outcome. Appendix \ref{apd:AO3-Cohort} gives baseline characteristics.

\paragraph{Application 1: Prediction} 

Performance metrics are in Table \ref{tab:rds-results}. General trends from worst to best were \emph{KLR} and \emph{XGBoost}, \emph{LR}, \emph{HFSM-Seq}, and then \emph{HFSM-Sim}. For AUROC and AUPRC performance, \emph{HFSM-Seq} and \emph{HFSM-Sim} were best. Calibration was best for \emph{HFSM-Sim}; all models tended to overestimate risk. \emph{LR} and \emph{LR-E} performed similarly on all metrics. No model demonstrated large performance gains over all others. Kernel containing models were slowest to train but still feasible to run. 

The \emph{J-CR} kernel was selected over \emph{J} three times for \emph{HFSM-Seq} and \emph{KLR} and two times for \emph{HFSM-Sim}. For all models, the combined provider and service type data were selected most often (Appendix \ref{apd:AO3-CaseStudyPred}).
\begin{table*}[!ht]
\centering
\caption{Clinical case study predictive performance results.}\label{tab:rds-results}
\begin{threeparttable}
\begin{tabular}{lllllll}
\toprule
  & LR & LR-E & KLR & HFSM-Seq & HFSM-Sim & XGBoost\\
\midrule
AUROC & 0.753 & 0.754 & 0.734 & 0.774 & 0.778 & 0.727\\
AUPRC & 0.146 & 0.148 & 0.139 & 0.185 & 0.184 & 0.137\\
Calibration Slope & 0.852 & 0.848 & 0.698 & 0.788 & 0.875 & 0.868\\
Calibration Intercept & -0.367 & -0.378 & -0.788 & -0.521 & -0.294 & -0.621\\
Time (minutes) & $<$ 1 & $<$ 1 & 42 & 115 & 89 & $<$ 1\\
\bottomrule
\end{tabular}
\footnotesize{\emph{Note:} Results were averaged across the five outer folds. \emph{Legend}: LR = Logistic Regression; LR-E = Logistic Regression-Extra Clinical; KLR = Kernel Logistic Regression; HFSM-Seq = Hybrid Feature- and Similarity-based Model-Sequential; HFSM-Sim = HFSM-Simultaneous.}
\end{threeparttable}
\end{table*}

\paragraph{Application 2: Interpretation} 

In general, \emph{HFSM-Seq} feature coefficients (Appendix \ref{apd:AO3-CaseStudyInterp}) were larger in magnitude and consistent in direction with those of \emph{HFSM-Sim}, suggesting the kernel adjusted most feature coefficients. We suspect this is a mix of the kernel serving as a mediator (e.g., stable housing) and confounder (e.g., depression or anxiety). This or collider bias could explain feature coefficients that increased (e.g., language) or qualitative change in sign (e.g., food insecurity). Refinement of the causal structure of the model would be needed to reliably estimate causal effects on risk or before deploying it to inform clinical decision making.


In \emph{HFSM-Seq}, $\alpha$s ranged from -1.70 to 1.30 and when rounded to five significant digits there were 5,038 zero, 13 positive, and 19 negative. Feature and outcome values stratified across these three groups are in Appendix \ref{apd:AO3-CaseStudyInterp}. No clients with negative $\alpha$ had the outcome, all lived in an urban geography, and obesity tended to be more prevalent than in other strata. Clients with positive $\alpha$ all spoke English as their primary language and tended to have lower household income and higher levels of stable housing, substance use, smoking or tobacco use, and food insecurity. 

The top ten weighted codes from each NMF topic for the three subgroups are in Appendix \ref{apd:AO3-CaseStudyInterp}. Clients with negative $\alpha$s had unique topics characterized by diagnosis and treatment with physician and nurse providers; related to counselling and foot care with counsellor and chiropodists; and related to counselling with nurse practitioners. The group with positive $\alpha$ had a unique topic strongly characterized by external referral and consult; a topic strongly characterized by social worker, nurse practitioner, and individual counselling; and overall less prominently featured codes related to diagnosis, treatment, and management. For zero $\alpha$, there was one topic strongly characterized by community resources and community health workers, which only weakly entered topics for the other subgroups. 


\section{Discussion}\label{sec:Discussion} 

We developed \emph{HFSM} for incorporating features where the outcome can be specified in a linear model (e.g., known informative risk factors or structured one-time question fields), as well as more complex data (e.g., historical data on care and diagnoses received) that are more easily incorporated as similarity measures. Our clinical case study demonstrated how \emph{HFSM} can be used for prediction and for exploring risk drivers.

\paragraph{Hybrid model methodology.} The hybrid models are most advantageous when both feature- and kernel-based data are important for the outcome. 
The predictive performance of \emph{HFSM-Sim} will be as good or better than \emph{HFSM-Seq}, assuming appropriate set up and tuning. \emph{HFSM-Seq} feature coefficients are interpreted analogously to \emph{LR}, while allowing the kernel to improve risk predictions. \emph{HFSM-Sim} feature coefficients are additionally adjusted for the kernel, which is valuable for certain clinical epidemiology tasks (e.g., complex confounder adjustment). The \emph{HFSM} form also allows for some or all feature coefficients to be fixed based on previous research studies or epidemiological analyses. Careful modelling is needed for explicit causal inference and when a risk prediction model might be interpreted as an ``upstream" treatment effect model. For example, if feature coefficients are interpreted as identifying modifiable risk factors (e.g., hypertension, smoking) to inform risk prevention strategy selection. Future work is needed to determine what ''pragmatic" level of causality is sufficient to support decisions in these settings.

To guide model development for interpretation tasks, frameworks such as directed acyclic graphs \citep{tennant2021}, and/or advanced techniques for multiple causal inference, such as the deconfounder approach \citep{wang2018, wang2019, greenland1999}, may be useful. Avenues for future work include interactions between features and kernel(s), other outcomes types, multilevel modelling, and adding an L2 penalty to kernel coefficients. The most closely related work to the latter applies an elastic net penalty to the dual form of the problem only \citep{feng2016}. 



\paragraph{Clinical case study.} For predicting the rare outcome of social isolation or loneliness in a complex primary health care population the \emph{HFSM} models performed as well or better than solely feature- or similarity-based models, including \texttt{XGboost}, while providing superior interpretability. Our proposed kernel \emph{J-CR} was selected more often than \emph{J} and may be useful in other contexts where two observations that deviate from population-level expectations are more similar than if they fit the expected profile. Of note, the kernel data included provider types; some, e.g., social workers, may have a higher index of suspicion for the outcome, but only baseline data were used and all providers could code the outcome. Future work could devise different cut-points, possibly based on the training data like \citep{belanche2013}, and optimize the switching point between presence and absence coding. We will also work with the Alliance to move towards model development for the purpose of eventual implementation.



In addition to insights about \emph{HFSM} and the kernel, our clinical case study demonstrates insights relevant to future primary health care machine learning applications. In contrast to settings where care is initiated due to a problem (e.g., cancer diagnosis, emergency room visit), primary health care is sought out during all stages of health and for a wide variety of concerns across the lifecourse, there is variability in visit patterns within and between clients, and risk patterns change across across care history due to cumulative and acute factors \citep{worldhealthorganization2021a, kueper2022a}. 

\paragraph{Limitations}
Outcome recording was not blinded and can only be considered a proxy. Some patient characteristics were not time-stamped, and about 13\% of clients excluded for having less than three years of observation had their first event before 2017.

We identified three risk factors in the literature, Sensory Disability, Social Phobia, and Dementia or Alzheimer's Disease, that were rare ($<$ 1\% prevalence) in our baseline cohort data. The standardized morbidity ratios (SMR), representing the ratio of observed to expected number of outcome cases based on sex-specific rates in the remaining eligible population, showed higher than expected risk in each rare feature sub-population (Sensory Disability SMR = 2.44; Social Phobia SMR = 3.27, Dementia or Alzheimer's Disease SMR = 2.11). There were too few clients with these conditions to meaningfully quantify risk, and we could not find an explainable AI framework that addressed this scenario. Qualitative research may provide more actionable information within these sub-populations.


%

We did not provide confidence intervals particularly in the case of \emph{HFSM-Sim} because although the objective function is convex, the L1 penalty is non-smooth; future work could adapt techniques like the $m$-out-of-$n$ bootstrap \citep{bickel2008} or use a selective inference framework \citep{chen2022} to account for the resulting non-regularity in the estimators.


\section{Conclusion}
\label{sec:conclusion}

Our work combines the perspectives and techniques of prediction-focused machine learning with interpretation-focused clinical epidemiology by supporting the use of complex data while maintaining interpretability and reproducibility. We demonstrated the utility of \emph{HFSM} by creating risk models and exploring factors associated with social isolation and loneliness in a complex primary health care population, using our new kernels based on population-level deviations. We look forward to ongoing useful developments that cross these disciplines.




\bibliography{ml4h22_refs.bib}

\clearpage



\appendix

\section{Extended Technical Background}
\label{appendix:AO3-bkgd} 


Our article addresses supervised learning tasks where the outcome of
interest ${y}(o)$ for a particular observation $o$ can be explained
partly by constructed features $\vec{\phi}(o)$ (i.e.\ scalar or fixed-length
vectors that represent a characteristic or property of $o$) and partly
by more complex information $\Psi(o)$ (e.g., high-dimensional,
time-varying, variable-length data). For example, in a primary health
care setting each observation could be a client and the outcome of
interest a condition or situation, such as diabetes or food insecurity,
that the client is at risk for and early intervention may help to
prevent. In this setting, $\phi(o)$-type information may include
sociodemographic characteristics and current diagnoses while
$\Psi(o)$-type information may include years of encounter data
representing the subset of thousands of possible tests, diagnoses, and
procedures that the client has received in their lifetime.

\subsection{Feature-based learning}

Feature-based supervised models, such as trees, learn a mathematical
function that takes features as input and provides an estimate about the
outcome as output \citep{bishop2006, russell2010}. To use a feature-based
approach for the above scenario, the $\Psi(o)$-type information must be
converted into $\phi(o)$-type information. This can be done in a
data-driven way and/or based on medical or social theories of health. For example, a data-driven dimensionality reduction approach such as topic modelling could be applied to client histories of diagnostic codes and the resulting topic weights for each client used as features. A theoretical approach may include identifying known risk-factors for the outcome based on research literature or clinical expertise, and then collapsing specific subsets of codes to generate binary indicators for whether or not the client has ever experienced each risk factor. Oftentimes feature construction loses or misses information and in general the extent to which this is a disadvantage will depend on the complexity of the data and predictive task.

In this article we used logistic regression (LR) to relate $\phi(o)$-type
information to the outcome. A prediction for a client of interest $o$
based on a set of $j$ features can be written
$\hat{y}(o)=\varsigma\left(\sum_{j}\phi_{j}(o)\beta_{j}\right)$ where
$\varsigma$ is the sigmoid function \citep{bishop2006, russell2010}.

\subsection{Similarity-based learning}

Similarity-based approaches, such nearest-neighbour and kernel methods, can handle both $\phi(o)$-type and $\Psi(o)$-type information as inputs \citep{bishop2006, russell2010}. Instead of learning explicit relationships between individual inputs and the outcome, these methods
use similarities or distances between observations by assuming that similar $o$ are likely to experience the same outcome, and if the same $o$ is entered into the model twice the same prediction will result. There are potentially two challenging aspects of $\Psi(o)$ that make similarity-based approaches attractive: the dimensionality and the proper form. In some situations, the best form for $\Psi(o)$ is known but its dimension is too large or challenging to construct with traditional feature-based approaches; other times, even if $\Psi(o)$ is a manageable size, the most useful way to incorporate it into a model is unknown.

Kernels can handle both of these challenges and are the similarity-based approach used in the remainder of this Chapter. A kernel function $k: \mathbb{R}^{m} \times \mathbb{R}^{m} \mapsto \mathbb{R}$ expresses the inner product between two inputs that have been mapped to
some high-dimensional feature space \citep{bishop2006, russell2010}. The feature mapping defines the notion of similarity captured by the scalar output, and can be non-linear with infinite dimensions; the mapping does
not need to be made explicit to use a kernel function. A valid \emph{kernel function} must be symmetric and result in a positive semi-definite kernel matrix. The notion(s) of similarity to capture, and whether data pre-processing is warranted, will depend on the specific scenario. 

\paragraph{Select kernel functions}

The simplest kernel function is the \emph{linear kernel},
$k(o_{i},o_{j}) = o_{i}^To_{j}$, which can be used to create a dual
formulation of linear regression that uses $o$ directly as features.

An example of a more complex and commonly used kernel function is the
\emph{Gaussian or Radial Basis Function (RBF)},
$k(o_{i}, o_{j}) = e ^ {- \frac{||o_{i} - o_{j}|| _{2}^{2}} {2\sigma^{2}}}$, which has a feature space with infinite dimensions and one
hyperparameter $\sigma$ to be tuned \citep{bishop2006}. The RBF kernel is used in our simulation study experiments.

The \emph{Jaccard kernel function}, described here, and proposed extension, described in the article, is used in our clinical case study. The Jaccard similarity between two sets of codes, A and B, is
$J(A,B) = \frac{ (A\cap B) }{ (A \cup B)}$. The highest similarity is
when all codes in A are in B and vice versa, and the lowest similarity
is when no codes that are in A are also in B, regardless of the number
of codes. The relative prevalence of codes does not matter, which leads to potentially limiting characteristics in situations with a large number of possible codes. For example, when trying to capture similarity with the Electronic Nomenclature and Classification Of Disorders and Encounters for Family Medicine (ENCODE-FM) \citep{bernstein2009} vocabulary that includes over 4,000 unique codes to record care activities. Strengths and limitations of the Jaccard are introduced below through simple examples:  

Imagine a scenario where a ``model" client comes in for a blood test, receives a diagnosis of hypertension, and then comes in for a follow-up appointment that includes a prescription treatment renewal. As shown in Table \ref{tab:jacEx1pt}, there are three other clients that have the same diagnosis, but different surrounding care. The
Jaccard similarity (Table \ref{tab:jacEx1sim}) appears to work well in that:

\begin{itemize}
    \item Client 1 similarity with others is proportional to the number of
    codes the other clients have.
    \item Clients 2 and 3 have comparable similarity profiles.
\end{itemize}

\begin{table}[!htb]
    \caption{Jaccard similarity score example.}
    \centering
      \subcaption{Client care profiles}
\begin{tabular}{cccc}
\toprule
Client & Lab Test & Dx & Tx\\
\midrule
C1 & TRUE & TRUE & TRUE\\
C2 & TRUE & TRUE & FALSE\\
C3 & FALSE & TRUE & TRUE\\
C4 & FALSE & TRUE & FALSE\\
\bottomrule
\end{tabular} \label{tab:jacEx1pt}
      \centering
        \subcaption{Jaccard similarity} 
\begin{tabular}{lcccc}
\toprule
  & C1 & C2 & C3 & C4\\
\midrule
C1 & 1.00 & 0.67 & 0.67 & 0.33\\
C2 & 0.67 & 1.00 & 0.33 & 0.50\\
C3 & 0.67 & 0.33 & 1.00 & 0.50\\
C4 & 0.33 & 0.50 & 0.50 & 1.00\\
\bottomrule
\end{tabular} \label{tab:jacEx1sim}
\end{table}

Now imagine there is another code, ``Com", that is so common everyone has it: similarity scores all increase, but the ratio in terms of who is most similar to who else stays the same. While this type of universally present code is not overly concerning for predictive purposes, a limitation emerges when we imagine what happens to the similarity score
for two clients that \emph{do not have a very common code} (Table \ref{tab:jacEx1ptCom,tab:jacEx1simCom}): 

\begin{itemize}
    \item Their similarity with everyone else decreases. (This is desirable.) 
    \item Their similarity with each other does not increase even though they share an ``abnormality" from the population expectation. (We hypothesize that this is not desirable.)
\end{itemize}

\begin{table}[h]
        \caption{Jaccard similarity score example with common code.}
        \centering
        \subcaption{Client care profiles}\label{tab:jacEx1ptCom} 
\begin{tabular}{ccccc}
\toprule
Client & Lab Test & Dx & Tx & Com\\
\midrule
C1 & TRUE & TRUE & TRUE & TRUE\\
C2 & TRUE & TRUE & FALSE & TRUE\\
C3 & FALSE & TRUE & TRUE & TRUE\\
C4 & FALSE & TRUE & FALSE & TRUE\\
C3b & FALSE & TRUE & TRUE & FALSE\\
C4b & FALSE & TRUE & FALSE & FALSE\\
\bottomrule
\end{tabular} \bigskip
      \subcaption{Jaccard similarity }\label{tab:jacEx1simCom} 
\begin{tabular}{lcccccc}
\toprule
  & C1 & C2 & C3 & C4 & C3b & C4b\\
\midrule
C1 & 1.00 & 0.75 & 0.75 & 0.50 & 0.50 & 0.25\\
C2 & 0.75 & 1.00 & 0.50 & 0.67 & 0.25 & 0.33\\
C3 & 0.75 & 0.50 & 1.00 & 0.67 & 0.67 & 0.33\\
C4 & 0.50 & 0.67 & 0.67 & 1.00 & 0.33 & 0.50\\
C3b & 0.50 & 0.25 & 0.67 & 0.33 & 1.00 & 0.50\\
C4b & 0.25 & 0.33 & 0.33 & 0.50 & 0.50 & 1.00\\
\bottomrule
\end{tabular} \end{table}

Sharing in the absence of common codes is not explicitly worked into the Jaccard similarity score. There are situations where the absence of common codes may matter more than the presence, for example missing check-ups after the diagnosis of a new condition or not receiving screening tests (when eligible). The same is not true for rare codes, where the presence of codes is generally expected to be more important than the absence. So, simply reverse coding everything or removing codes from the universe is not a suitable solution. A modification to the Jaccard function to account for limitations is presented in our article.

\section{Illustrative Example: Primal and Dual Forms}\label{apd:AO3-PDexample} 

The feature- and similarity-based parts of the model (\emph{HFSM}) can represent primal and dual forms, respectively \citep{bishop2006}. Thus, a model where some features are combined using an unpenalized linear kernel will be equivalent to a model where all features are entered in logistic regression. To demonstrate this, we generated 10,000 observations according to the data generating mechanism $P(Y) = \sigma(0.25-1X_{1} + 2X_{2})$ where $X_{1,2} \sim N(0,1)$. Logistic regression was fit with an intercept, $X_{1}$, and $X_{2}$; \emph{HFSM} was fit with the intercept and $X_{1}$ maintained as features and $X_{2}$ included with a linear kernel.

\begin{table}[hbp]
\caption{\label{tab:ao3-illustrative-ex-table}Illustrative example feature coefficients}
\centering
\begin{tabular}[t]{lcc}
\toprule
  & LR & HFSM\\
\midrule
$\beta_0$ & 0.24 & 0.24\\
$\beta_1$ & -1.04 & -1.04\\
$\beta_2$ & 2.04 & NA\\
\bottomrule
\end{tabular}
\end{table}

Table \ref{tab:ao3-illustrative-ex-table} shows the learned coefficients whereby the intercept and $X_{1}$ coefficients were the same for the two models. The unpenalized $\alpha$'s from \emph{HFSM} ranged from -0.001 to 0.001. As expected, predictions based on the two model forms were also equivalent (not shown). 

\clearpage

\onecolumn 

\section{Hybrid Model Code}\label{apd:AO3-modelcode}

The following Python code can be used to fit the four main models we used: logistic regression (M1), kernel logistic regression (M2), hybrid model sequential fit (M3), and hybrid model simultaneous fit (M4).

\lstset{basicstyle=\footnotesize\ttfamily,language=Python}
\begin{lstlisting}

import numpy as np
import cvxpy as cp
from sklearn.metrics import roc_auc_score
from scipy.special import expit

# Function to fit feature only model
# @param Xtrain the training feature data matrix
# @param yTrain training binary outcome vector
# @return betas and auc on training data

# Note that cp.logistic(x) is log(1 + exp(x)), not sigmoid

def FIT_M1(Xtrain, yTrain, save=False, fnHead=None):

    beta = cp.Variable((Xtrain.shape[1], 1))
    
    problemM1 = cp.Problem(cp.Maximize(cp.sum(
                    cp.multiply(yTrain, (Xtrain @ beta))
                    - cp.logistic((Xtrain @ beta)))/Xtrain.shape[0]))

    problemM1.solve(verbose=False, solver=cp.ECOS)

    # Get training AUC value
    aucTrain = roc_auc_score(yTrain, expit(Xtrain @ beta.value))

    print(f"\n *** DONE M1 FIT ***"
          f"\nStatus of M1 problem: {problemM1.status}"
          f"and Optimal value: {problemM1.value}"
          f"and solve time: {problemM1._solve_time}"
          f"\n**Training AUC: {aucTrain}"
          )

    if (save):
        np.save(fnHead + "_Betas.npy", beta.value)
        np.save(fnHead + "_SolveTime.npy", problemM1._solve_time)
        np.save(fnHead + "_OptValue.npy", problemM1.value)
        np.save(fnHead + "_Status.npy", problemM1.status)
        np.save(fnHead + "_aucTrain.npy", aucTrain)

    return beta.value, aucTrain




# Function to fit kernel only model with L1 penalty
# @param Ktrain precomputed training kernel matrix
# @param yTrain training outcome vector
# @param l1 strength of L1 penalty for alphas
# @param fnHead start path to save object
#  including directory and foldO
# @return alphas, auc on training data
def FIT_M2(Ktrain, yTrain, l1, save=False, fnHead=None):

    alpha = cp.Variable((Ktrain.shape[1], 1))

    lam = cp.Parameter(nonneg=True, value=l1)

    problemM2 = cp.Problem(cp.Maximize(cp.sum(
                    cp.multiply(yTrain, (Ktrain @ alpha))
                    - cp.logistic(Ktrain @ alpha))/Ktrain.shape[0]
                    - lam * cp.norm(alpha, 1)))

    problemM2.solve(verbose=False, solver=cp.ECOS)

    # Get training AUC value
    aucTrain = roc_auc_score(yTrain, expit(Ktrain @ alpha.value))

    print(f"\n *** DONE M2 FIT WITH LAM {l1} ***"
          f"\nStatus of M2 problem: {problemM2.status}"
          f"and Optimal value: {problemM2.value}"
          f"and solve time: {problemM2._solve_time}"
          f"\n**Training AUC: {aucTrain}"
          )

    if (save):
        np.save(fnHead + "_Alphas.npy", alpha.value)
        np.save(fnHead + "_SolveTime.npy", problemM2._solve_time)
        np.save(fnHead + "_OptimalValue.npy", problemM2.value)
        np.save(fnHead + "_Status.npy", problemM2.status)
        np.save(fnHead + "_aucTrain.npy", aucTrain)

    return alpha.value, aucTrain


# Function to fit HFSM-Seq with L1 penalty
# Betas are fit first and fixed while learning alphas
# @param Xtrain precomputed training feature matrix
# @param Ktrain precomputed training kernel matrix
# @param yTrain training outcome vector
# @param l1 strength of L1 penalty for alphas
# @param fnHead start path to save object
#  including directory and foldO
# @return betas, alphas, auc on training data
def FIT_M3(Xtrain, Ktrain, yTrain, l1, fixedBeta=None, save=False, fnHead=None):

    if (fixedBeta==None):
        # learn the betas ignoring alphas
        fixedBeta, aucNotUsed = FIT_M1(Xtrain, yTrain,
                                        save=True, fnHead=fnHead + "_m1Part")

    # betas are set up as fixed parameter for learning alphas
    betaM1 = cp.Parameter(fixedBeta.shape, value=fixedBeta)
    # Alphas are learned
    alpha = cp.Variable((Ktrain.shape[0], 1))
    # L1 penalty strength is fixed parameter
    lam = cp.Parameter(nonneg=True, value=l1)

    # problem to solve
    problemM3 = cp.Problem(cp.Maximize(cp.sum(
                        cp.multiply(yTrain, (Ktrain @ alpha + Xtrain @ betaM1))
                        - cp.logistic(Ktrain @ alpha + Xtrain @ betaM1))
                        / Ktrain.shape[0]
                        - lam * cp.norm(alpha, 1)))

    # call the solver; default max iters is 10,000
    problemM3.solve(verbose=False, warm_start=True, solver=cp.ECOS)

    aucTrain = roc_auc_score(yTrain,
                             expit(Xtrain @ betaM1.value + Ktrain @ alpha.value))

    print(f"\n *** DONE M3 FIT WITH LAM {l1} ***"
          f"\nStatus of problem: {problemM3.status}"
          f"and Optimal value: {problemM3.value}"
          f"and solve time: {problemM3._solve_time}"
          f"\n**Training AUC: {aucTrain}"
          )

    if (save):
        np.save(fnHead + "_BetasM1.npy", betaM1.value) 
        np.save(fnHead + "_Alphas.npy", alpha.value)
        np.save(fnHead + "_SolveTime.npy", problemM3._solve_time)
        np.save(fnHead + "_OptimalValue.npy", problemM3.value)
        np.save(fnHead + "_Status.npy", problemM3.status)
        np.save(fnHead + "_aucTrain.npy", aucTrain)

    return betaM1.value, alpha.value, aucTrain


# Function to fit HFSM-Sim with L1 penalty
# @param Xtrain precomputed training feature matrix
# @param Ktrain precomputed training kernel matrix
# @param yTrain training outcome vector
# @param l1 strength of L1 penalty for alphas
# @param fnHead start path to save object
#  including directory and foldO
# @return betas, alphas, auc on training data data
def FIT_M4(Xtrain, Ktrain, yTrain, l1, save=False, fnHead=None):

    # Variables can be scalars, vectors, or matrices
    beta = cp.Variable((Xtrain.shape[1], 1))
    # vector of values (n,1) to fit
    alpha = cp.Variable((Ktrain.shape[0], 1))
    # Parameter - this one is positive scalar for lam
    lam = cp.Parameter(nonneg=True, value=l1)

    # problem to solve
    problemM4 = cp.Problem(cp.Maximize(cp.sum(
                        cp.multiply(yTrain, (Ktrain @ alpha + Xtrain @ beta))
                        - cp.logistic(Ktrain @ alpha + Xtrain @ beta))
                        / Ktrain.shape[0]
                        - lam * cp.norm(alpha, 1)))

    # call the solver; default max iters is 10,000
    problemM4.solve(verbose=False, warm_start=True, solver=cp.ECOS)

    aucTrain = roc_auc_score(yTrain,
                             expit(Xtrain @ beta.value + Ktrain @ alpha.value))

    print(f"\n*** DONE M4 FIT WITH LAM {l1} ***"
          f"\nStatus of problem: {problemM4.status}"
          f"and Optimal value: {problemM4.value}"
          f"and solve time: {problemM4._solve_time}"
          f"\n**Training AUC: {aucTrain}"
          )

    if (save):
        np.save(fnHead + "_Betas.npy", beta.value)
        np.save(fnHead + "_Alphas.npy", alpha.value)
        np.save(fnHead + "_SolveTime.npy", problemM4._solve_time)
        np.save(fnHead + "_OptimalValue.npy", problemM4.value)
        np.save(fnHead + "_Status.npy", problemM4.status)
        np.save(fnHead + "_aucTrain.npy", aucTrain)

    return beta.value, alpha.value, aucTrain


\end{lstlisting}

\clearpage

\section{Illustrative Examples: Interpretation}\label{apd:interpEx} 

\begin{figure}[H]
\parbox{\columnwidth}{
  \parbox{0.24\columnwidth}{
    \centering
    \includegraphics[width=0.24\textwidth]{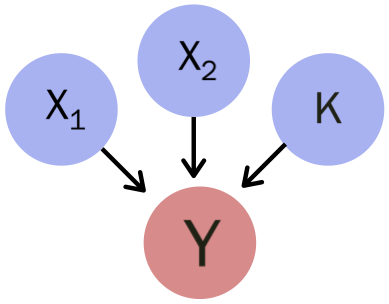}
    Example 1
  }
  \parbox{0.24\columnwidth}{
    \centering
    \includegraphics[width=0.24\textwidth]{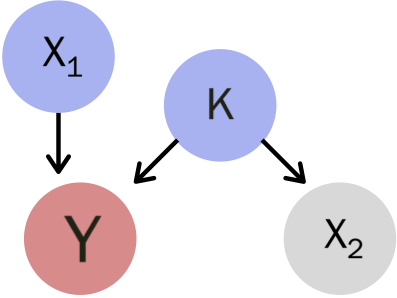}
    Example 2
  }
  \parbox{0.24\columnwidth}{
    \centering
    \includegraphics[width=0.18\textwidth]{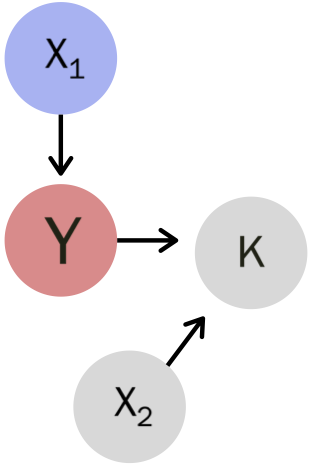}
    Example 3
}
\parbox{0.24\columnwidth}{
    \centering
    \includegraphics[width=0.24\textwidth]{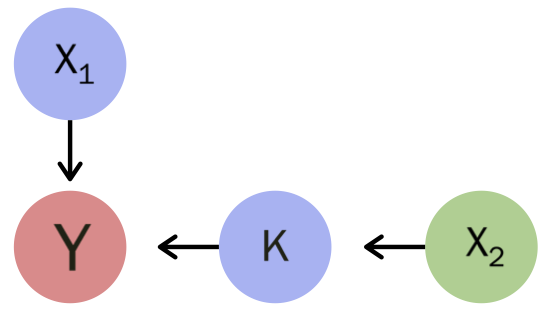}
    Example 4
}
}
\caption{Data generating mechanisms used to contrast sequential and simultaneous hybrid model optimization.}%
\label{fig:sInterp}%
\end{figure}

We set up a series of illustrative examples contrast the performance and interpretation of \emph{HFSM-Seq} and \emph{HFSM-Sim} in terms of causal inference. For each example there was a binary outcome $y$, one continuous feature $X_{1} \sim N(0,1)$ that maintained a direct relationship with $y$, and a binary feature $X_{2}$ whose relationship with $y$ and $K$ was manipulated. For simplicity, $K$ was unpenalized and constructed from a linear kernel function applied to a single binary variable. We designed four examples, represented in Figure \ref{fig:sInterp}:

\begin{enumerate}[noitemsep, topsep=0pt]
  \item \textbf{Independent contributions}. $P(Y) = \varsigma(0.25 -1X_{1} + 2X_{2} + 3K )$ where $X_{2} \sim B(p=0.5)$, and $K \sim B(p=0.5)$. 
  \item The kernel operated as a \textbf{confounder} between the second feature and the outcome. $P(Y) = \varsigma(0.25 -1X_{1} + 3K)$ where $K \sim B(p=0.5)$ and $P(X_{2}) = \varsigma(2 K)$. 
  \item The kernel operated as a \textbf{collider} between the outcome and on the second feature. $P(Y) = \varsigma(0.25 -1X_{1})$ where $P(K) = \varsigma(3Y + 2X_{2})$ and $X_{2} \sim B(p=0.5)$. 
  \item The kernel operated as a \textbf{mediator} between the second feature and the outcome. $P(Y) = \varsigma(0.25 -1X_{1} + 3K)$ where $P(K) = \varsigma(2X_{2})$ and $X_{2} \sim B(p=0.5)$.
\end{enumerate}

For each example, feature coefficients were compared for \emph{HFSM-Seq}, \emph{HFSM-Sim}, and \emph{LR} fit on 3,000 training examples and predictive performance was compared based on area under the receiver operating characteristic curve (AUROC) for 1,000 new test examples. Results are presented in Table \ref{tab:illustrativeExInterp}.  

\begin{table}[htp]
\caption{AUROCs and coefficients for interpretation example.} \label{tab:illustrativeExInterp}
\begin{minipage}[t]{\textwidth}
\subcaption{Test AUROCs}
\centering
 \begin{tabular}[t]{lrrr}
\toprule
 & LR & HFSM-Seq & HFSM-Sim\\
\midrule
Ex. 1 & 0.758 & 0.860 & 0.863\\
Ex. 2 & 0.721 & 0.834 & 0.854\\
Ex. 3 & 0.777 & 0.800 & 0.845\\
Ex. 4 & 0.739 & 0.852 & 0.888\\
\bottomrule
\end{tabular}
\end{minipage}
\begin{minipage}[t]{\textwidth}
\subcaption{Feature coefficients} 
\centering
  \begin{tabular}[t]{lrrr}
\toprule
  & HFSM-Seq & HFSM-Sim & True\\
\midrule
\addlinespace[0.3em]
\multicolumn{4}{l}{\textbf{Ex. 1: Independent Contributions}}\\
\hspace{1em}$\beta_0$ & 1.345 & 0.395 & 0.25\\
\hspace{1em}$\beta_1$ & -0.796 & -0.995 & -1.00\\
\hspace{1em}$\beta_2$ & 1.515 & 1.896 & 2.00\\
\addlinespace[0.3em]
\multicolumn{4}{l}{\textbf{Ex. 2: K was a Confounder}}\\
\hspace{1em}$\beta_0$ & 0.595 & 0.264 & 0.25\\
\hspace{1em}$\beta_1$ & -0.723 & -0.960 & -1.00\\
\hspace{1em}$\beta_2$ & 1.044 & 0.092 & 0.00\\
\addlinespace[0.3em]
\multicolumn{4}{l}{\textbf{Ex. 3: K was a Collider}}\\
\hspace{1em}$\beta_0$ & 0.382 & -2.325 & 0.25\\
\hspace{1em}$\beta_1$ & -0.950 & -0.979 & -1.00\\
\hspace{1em}$\beta_2$ & -0.020 & -0.578 & 0.00\\
\addlinespace[0.3em]
\multicolumn{4}{l}{\textbf{Ex. 4: K was a Mediator}}\\
\hspace{1em}$\beta_0$ & 1.213 & 0.295 & 0.25\\
\hspace{1em}$\beta_1$ & -0.793 & -0.984 & -1.00\\
\hspace{1em}$\beta_2$ & 1.246 & 0.158 & 2.00\\
\bottomrule
\end{tabular}
\end{minipage}
 \footnotesize \raggedright \singlespacing {\emph{Note:} HFSM-Seq and logistic regression coefficients are equal. \emph{Legend}: HFSM-Seq = Hybrid Model - Sequential Fit, HFSM-Sim = Hybrid Model - Simultaneous Fit, LR = Logistic Regression, True = Coefficients used to generate the data.}
\end{table}


\onecolumn 
\section{Simulation Study Details}\label{apd:AO3-simulationstudy} 

\subsection{RBF kernel sigma selection}\label{apd:AO3-rbfSigSelect} 

Three candidate hyperparameter values for the RBF kernel were selected to provide a range of diagonal dominance as assessed by the following equation:

$$DD = \sum \frac{|\mathit{diagonals}|}{|\mathit{off-diagonals}|}$$ For
a matrix with $i,j=n$ observations:
$$DD = \sum_{i,j} \frac{|a_{i,i}|}{|a_{i,j}| - |a_{i,i}|}$$

For a kernel matrix $a_{i,i} = 1$ and the range will be
$\left[ \frac{n}{(n-1)}, \infty \right)$.\\  

\begin{figure}[h]
\parbox{\columnwidth}{
  \parbox{0.3\columnwidth}{
    \centering
    \includegraphics[width=0.3\textwidth]{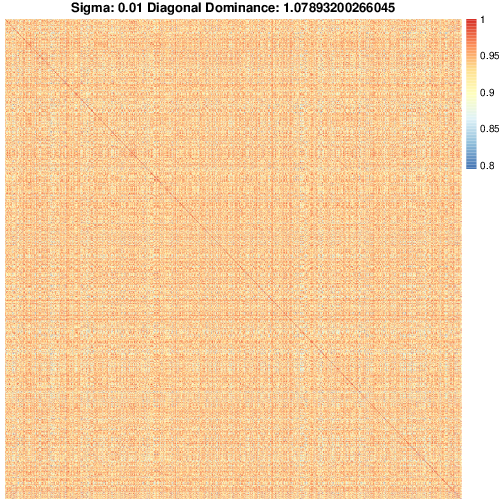}
    $\sigma = 0.01$
  }
  \parbox{0.3\columnwidth}{
    \centering
    \includegraphics[width=0.3\textwidth]{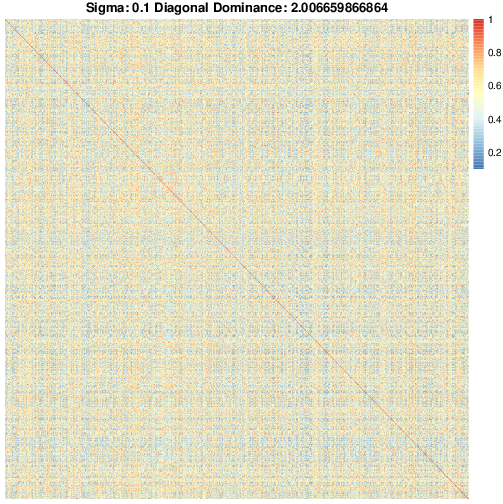}
    $\sigma = 0.1$
  }
  \parbox{0.3\columnwidth}{
    \centering
    \includegraphics[width=0.3\textwidth]{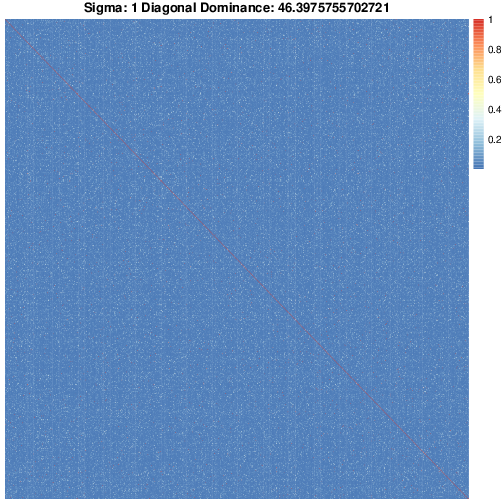}
    $\sigma = 1$
}
}
\caption{Heatmap demonstrating similarity of RBF kernel with various $\sigma$ on a random sample.}
\label{fig:rbfHeatmaps}
\end{figure}

In addition to looking at the above scalar measure, we generated heat plots for RBF kernels with a range of $\sigma$ values on a random sample of
1000 observations of the six variables in MONK's data problems. Example plots in Figure \ref{fig:rbfHeatmaps} provide another view at how varying the $\sigma$ values alters the similarity captured by the RBF kernel.

\onecolumn

\subsection{Selected hyperparameters and model coefficients}\label{apd:AO3-SSextraRes}  


\begin{table}[ht]

\caption{Synthetic data scenario 1: selected hyperparameters} \label{ao3-S1hyper}
\centering
\begin{tabular}[t]{lccccc}
\toprule
  & Fold 1 & Fold 2 & Fold 3 & Fold 4 & Fold 5\\
\midrule
\addlinespace[0.3em]
\multicolumn{6}{l}{\textbf{Sigma for RBF Kernel}}\\
\hspace{1em}LR & NA & NA & NA & NA & \vphantom{1} NA\\
\hspace{1em}KLR & 1.000 & 1.000 & 1.000 & 1.000 & 1.000\\
\hspace{1em}HFSM-Seq & 0.100 & 1.000 & 0.100 & 0.100 & 1.000\\
\hspace{1em}HFSM-Sim & 0.100 & 1.000 & 0.100 & 0.100 & 0.100\\
\addlinespace[0.3em]
\multicolumn{6}{l}{\textbf{L1 Penalty Strength}}\\
\hspace{1em}LR & NA & NA & NA & NA & NA\\
\hspace{1em}KLR & 0.032 & 0.001 & 0.001 & 1.000 & 1.000\\
\hspace{1em}HFSM-Seq & 0.001 & 0.001 & 0.001 & 0.001 & 0.001\\
\hspace{1em}HFSM-Sim & 0.001 & 0.001 & 0.001 & 0.001 & 0.001\\
\bottomrule
\end{tabular}

\end{table}

\begin{table}[ht]
\caption{Synthetic data scenario 1: model interpretation} \label{tab:ao3-S1interp}
\begin{minipage}[t]{0.5\textwidth}
\subcaption{Average Feature Coefficients}
\centering 

\begin{tabular}[t]{lccc}
\toprule
  & LR & HFSM-Sim & True\\
\midrule
$\beta_0$ & -1.263 & -1.021 & -1.500\\
$\beta_1$ & 0.320 & 0.319 & 0.300\\
$\beta_2$ & 0.440 & 0.444 & 0.400\\
$\beta_3$ & 0.568 & 0.567 & 0.600\\
$\beta_4$ & 0.695 & 0.695 & 0.700\\
\bottomrule
\end{tabular}
\end{minipage}
\begin{minipage}[t]{0.5\textwidth}
\subcaption{Average Kernel Coefficients}
\centering 
\begin{tabular}[t]{lccc}
\toprule
  & KLR & HFSM-Seq & HFSM-Sim\\
\midrule
Non-0 & 262.600 & 197.400 & 180.400\\
Max & 0.000 & 0.014 & 0.005\\
Min & -0.019 & -0.014 & -0.019\\
Mean & -0.004 & 0.000 & -0.005\\
Median & -0.003 & -0.002 & -0.006\\
\bottomrule
\end{tabular}
\end{minipage}
\end{table}


\begin{table}[ht]

\caption{Synthetic data scenario 2: selected hyperparameters} \label{ao3-S2hyper}
\centering
 \begin{tabular}[t]{lccccc}
\toprule
  & Fold 1 & Fold 2 & Fold 3 & Fold 4 & Fold 5\\
\midrule
\addlinespace[0.3em]
\multicolumn{6}{l}{\textbf{Sigma for RBF Kernel}}\\
\hspace{1em}LR & NA & NA & NA & NA & \vphantom{1} NA\\
\hspace{1em}KLR & 1.000 & 1.000 & 1.000 & 1.000 & 1.000\\
\hspace{1em}HFSM-Seq & 1.000 & 1.000 & 1.000 & 1.000 & 1.000\\
\hspace{1em}HFSM-Sim & 1.000 & 1.000 & 1.000 & 1.000 & 1.000\\
\addlinespace[0.3em]
\multicolumn{6}{l}{\textbf{L1 Penalty Strength}}\\
\hspace{1em}LR & NA & NA & NA & NA & NA\\
\hspace{1em}KLR & 0.001 & 0.001 & 0.001 & 0.001 & 0.001\\
\hspace{1em}HFSM-Seq & 0.001 & 0.001 & 0.001 & 0.001 & 0.001\\
\hspace{1em}HFSM-Sim & 0.001 & 0.001 & 0.001 & 0.001 & 0.001\\
\bottomrule
\end{tabular}
\end{table}

\begin{table}[ht]

\caption{Synthetic data scenario 2: model interpretation} \label{tab:ao3-s2interp}
\begin{minipage}[t]{0.5\textwidth}
\subcaption{Average Feature Coefficients}
\centering 
\begin{tabular}[t]{lccc}
\toprule
  & LR & HFSM-Sim & True\\
\midrule
$\beta_0$ & -0.844 & -0.557 & -2.100\\
$\beta_1$ & 0.185 & 0.198 & 0.300\\
$\beta_2$ & 0.377 & 0.408 & 0.400\\
$\beta_3$ & 0.443 & 0.457 & 0.600\\
$\beta_4$ & 0.566 & 0.605 & 0.700\\
\bottomrule
\end{tabular}
\end{minipage}
\begin{minipage}[t]{0.5\textwidth}
\subcaption{Average Kernel Coefficients} 
\centering
\begin{tabular}[t]{lccc}
\toprule
  & KLR & HFSM-Seq & HFSM-Sim\\
\midrule
Non-0 & 1696.200 & 1732.800 & 1631.000\\
Max & 0.082 & 0.085 & 0.082\\
Min & -0.098 & -0.096 & -0.109\\
Mean & -0.002 & -0.001 & -0.010\\
Median & 0.001 & 0.001 & -0.006\\
\bottomrule
\end{tabular}
\end{minipage}
\end{table}

 
\begin{table}[ht]

\caption{Synthetic data scenario 3: selected hyperparameters} \label{ao3-S3hyper}
\centering
\begin{tabular}[t]{lccccc}
\toprule
  & Fold 1 & Fold 2 & Fold 3 & Fold 4 & Fold 5\\
\midrule
\addlinespace[0.3em]
\multicolumn{6}{l}{\textbf{Sigma for RBF Kernel}}\\
\hspace{1em}LR & NA & NA & NA & NA & \vphantom{1} NA\\
\hspace{1em}KLR & 1.000 & 1.000 & 1.000 & 1.000 & 1.000\\
\hspace{1em}HFSM-Seq & 1.000 & 1.000 & 1.000 & 1.000 & 1.000\\
\hspace{1em}HFSM-Sim & 1.000 & 1.000 & 1.000 & 1.000 & 1.000\\
\addlinespace[0.3em]
\multicolumn{6}{l}{\textbf{L1 Penalty Strength}}\\
\hspace{1em}LR & NA & NA & NA & NA & NA\\
\hspace{1em}KLR & 0.001 & 0.001 & 0.001 & 0.001 & 0.001\\
\hspace{1em}HFSM-Seq & 0.001 & 0.001 & 0.001 & 0.001 & 0.001\\
\hspace{1em}HFSM-Sim & 0.001 & 0.001 & 0.001 & 0.001 & 0.001\\
\bottomrule
\end{tabular}
\end{table}

\begin{table}[ht]
\caption{Synthetic data scenario 3: model interpretation} \label{tab:ao3-s3interp}
\begin{minipage}[t]{0.5\textwidth}
\subcaption{Average Feature Coefficients}
\centering
\begin{tabular}[t]{lccc}
\toprule
  & LR & HFSM-Sim & True\\
\midrule
$\beta_0$ & -0.459 & 0.414 & -3.200\\
$\beta_1$ & 0.150 & 0.160 & 0.300\\
$\beta_2$ & 0.163 & 0.203 & 0.400\\
$\beta_3$ & 0.235 & 0.325 & 0.600\\
$\beta_4$ & 0.280 & 0.385 & 0.700\\
\bottomrule
\end{tabular}
\end{minipage}
\begin{minipage}[t]{0.5\textwidth}
\subcaption{Average Kernel Coefficients} 
\centering
\begin{tabular}[t]{lccc}
\toprule
  & KLR & HFSM-Seq & HFSM-Sim\\
\midrule
Non-0 & 2575.800 & 2599.600 & 2546.400\\
Max & 0.094 & 0.102 & 0.092\\
Min & -0.264 & -0.269 & -0.297\\
Mean & -0.004 & -0.003 & -0.021\\
Median & 0.006 & 0.006 & -0.013\\
\bottomrule
\end{tabular}
\end{minipage}
\end{table}

\clearpage

\section{Clinical Case Study Details}\label{ap:AO3-casestudy} 

\subsection{Cohort overview}\label{apd:AO3-Cohort}

\normalsize 

\paragraph{Cohort construction} Eligible clients included ongoing primary care clients at Urban-At-Risk (UAR) Community Health Centres (CHCs) without the outcome at baseline. We restricted to new or newly returning, mid- to long-term clients by requiring the first event record to be in 2010 or later and there to be at least one event three years from the first recorded event. Primary health care is provided at all stages of life and health and social isolation or loneliness may occur at any point, so we randomly selected two-year periods from each client's observation history to serve as the prediction interval. The start of the prediction interval had to be at least one year from the first recorded event as the first year of care provision in this population is associated with a distinct risk profile, likely due to ``catch-up" on unresolved care and diagnoses \citep{kueper2022a}. Feature and kernel input data were from the first recorded event to the beginning of the prediction interval. We restricted our cohort to those 45-64 years old at the end of their baseline period as age is associated with the outcome and may influence the risk factors and potential interventions to help someone at high-risk.

\paragraph{Loss to follow up} Of the 10,687 people excluded for having less than three years between their first and last care records, 6,276 (58.7\%) had their first event in 2017 or later so there was not enough calendar time for sufficient observation; bias due to their exclusions is expected to be minimal. The remaining 4,411 (41.3\%) were ``true" loss to follow-up under a more traditional research study paradigm; we do not know if they stopped receiving care altogether or if they switched to another health care organization. After applying additional eligibility criteria there were 1,430 clients and among them there were 108 cases of the outcome of which 22 (16.9\%) occurred at least one year from the first recorded event. If our study was more application than methods testing focused we would perform sensitivity analyses to assess whether there is bias due to these lost to follow up as in a real world setting the future length of care when applying a predictive model is unknown. 



\begin{figure}[H]
  \centering 
  \includegraphics[width=0.8\textwidth]{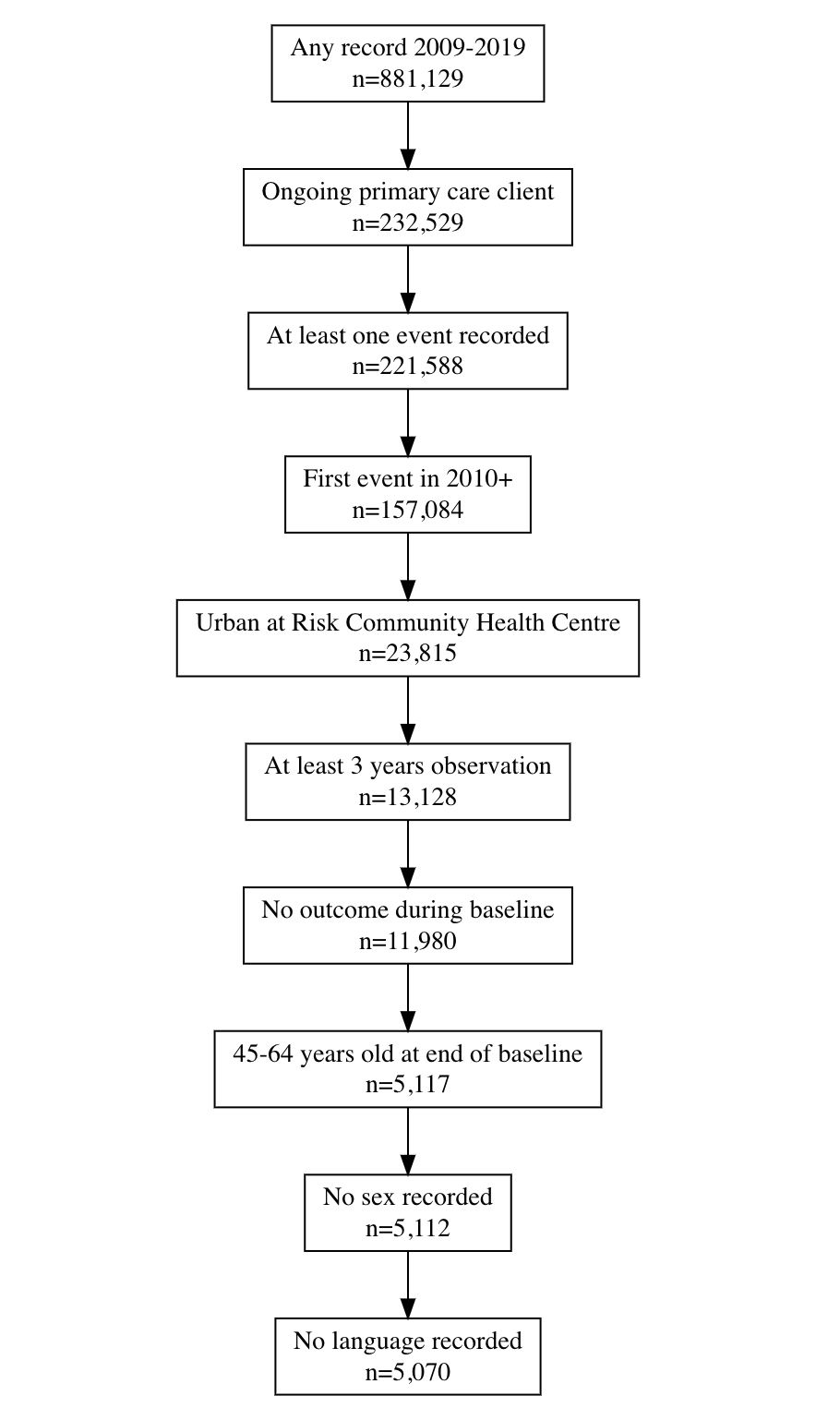}%
  \caption{Clinical case study cohort flow diagram}%
  \label{fig:ao3-flowDgm}%
\end{figure}

\begin{longtable}[t]{lll}
\caption{Clinical case study baseline features.}\\
\toprule
Feature & Values & n (\%)\\
\midrule
Sex & Female & 2379 (46.92)\\
 & Male & 2691 (53.08)\\
\cmidrule{1-3}\pagebreak[0]
Rural Residence & Rural & 1011 (19.94)\\
 & Urban & 3942 (77.75)\\
 & Missing & 117 (2.31)\\
\cmidrule{1-3}\pagebreak[0]
Household Income & \$0 to \$14,999 & 1254 (24.73)\\
 & \$15,000 to \$24,999 & 454 (8.95)\\
 & \$25,000 to \$34,999 & 274 (5.40)\\
 & \$35,000 to \$59,000 & 535 (10.55)\\
 & \$60,000 or more & 600 (11.83)\\
 & Do not know & 274 (5.40)\\
 & Prefer not to answer & 587 (11.58)\\
 & Missing & 1092 (21.54)\\
\cmidrule{1-3}\pagebreak[0]
Household Composition & Couple & 1897 (37.42)\\
 & Other Family & 519 (10.24)\\
 & Unrelated housemates & 217 (4.28)\\
 & Sole Member & 1205 (23.77)\\
 & Do Not Know or Other & 255 (5.03)\\
 & Prefer not to answer & 57 (1.12)\\
 & Missing & 920 (18.15)\\
\cmidrule{1-3}\pagebreak[0]
Education Completed & Post-secondary or equivalent & 1717 (33.87)\\
 & Secondary or equivalent & 1849 (36.47)\\
 & Less than high school & 395 (7.79)\\
 & Do Not Know or Other & 269 (5.31)\\
 & Prefer not to answer & 54 (1.07)\\
 & Missing & 786 (15.50)\\
\cmidrule{1-3}\pagebreak[0]
Language & English & 4691 (92.52)\\
 & French & 82 (1.62)\\
 & Other & 297 (5.86)\\
\cmidrule{1-3}\pagebreak[0]
LGBTQ & Lgbtq & 67 (1.32)\\
 & Non-Lgbtq & 1084 (21.38)\\
 & Missing & 3919 (77.30)\\
\cmidrule{1-3}\pagebreak[0]
Years in Canada & True & 627 (12.37)\\
\cmidrule{1-3}\pagebreak[0]
Physical Disability & True & 240 (4.73)\\
\cmidrule{1-3}\pagebreak[0]
Depression or Anxiety & True & 410 (8.09)\\
\cmidrule{1-3}\pagebreak[0]
Chronic Urinary Problem & True & 852 (16.80)\\
\cmidrule{1-3}\pagebreak[0]
Obesity & True & 737 (14.54)\\
\cmidrule{1-3}\pagebreak[0]
Personality Disorder & True & 145 (2.86)\\
\cmidrule{1-3}\pagebreak[0]
Stable Housing & True & 556 (10.97)\\
\cmidrule{1-3}\pagebreak[0]
Substance Use & True & 753 (14.85)\\
\cmidrule{1-3}\pagebreak[0]
Smoking or Tobacco Use & True & 1454 (28.68)\\
\cmidrule{1-3}\pagebreak[0]
Food Insecurity & True & 200 (3.94)\\
\cmidrule{1-3}\pagebreak[0]
\label{tab:ao3-rdsClinicalChara}
\end{longtable}

\normalsize


\subsection{Application 1: Prediction}\label{apd:AO3-CaseStudyPred} 

\begin{table}[htp]
\caption{Clinical case study selected hyperparameters.}
\centering
\begin{tabular}[t]{llllll}
\toprule
  & Fold 1 & Fold 2 & Fold 3 & Fold 4 & Fold 5\\
\midrule
\addlinespace[0.3em]
\multicolumn{6}{l}{\textbf{L1 Penalty Strength}}\\
\hspace{1em}KLR & 1e-04 & 1e-03 & 1e-04 & 1e-04 & 1e-04\\
\hspace{1em}HFSM-Seq & 1e-04 & 1e-03 & 1e-03 & 1e-03 & 1e-04\\
\hspace{1em}HFSM-Sim & 1e-03 & 1e-03 & 1e-03 & 1e-03 & 1e-04\\
\addlinespace[0.3em]
\multicolumn{6}{l}{\textbf{Kernel Function \& Data}}\\
\hspace{1em}KLR & J-CR\_PIST2 & J\_ST & J\_PIST2 & J-CR\_PIST2 & J-CR\_PIST2\\
\hspace{1em}HFSM-Seq & J-CR\_ST & J-CR\_PIST2 & J\_ST & J-CR\_PIST2 & J\_PIST2\\
\hspace{1em}HFSM-Sim & J\_ST & J-CR\_PIST2 & J\_PIST2 & J-CR\_PIST2 & J\_PIST2\\
\bottomrule
\multicolumn{6}{l}{\footnotesize \emph{Legend}: J = Jaccard similarity; J-CR = Jaccard - Common absence and rare presence similarity extension;}\\ 
\multicolumn{6}{l}{\footnotesize PI = Provider type data; ST = Service type data; PIST2 = both.}\\
\end{tabular}
\label{tab:ao3-clinicalSelectedHyp}
\end{table}


\subsection{Application 2: Inference/interpretation}\label{apd:AO3-CaseStudyInterp}

\begin{longtable}[t]{llcc}
\caption{Feature coefficients for models re-trained on all data}\\
\toprule
Variable & Values & HFSM-Seq & HFSM-Sim\\
\midrule
Intercept &  & -4.275 & -5.268\\
\cmidrule{1-4}\pagebreak[0]
Sex & Male & -0.261 & -0.202\\
\cmidrule{1-4}\pagebreak[0]
Rural Residence & Urban & 0.462 & 0.227\\
 & Missing & 0.774 & 0.636\\
\cmidrule{1-4}\pagebreak[0]
Household Income & \$15,000 to \$24,999 & 0.137 & 0.036\\
 & \$25,000 to \$34,999 & -0.492 & -0.444\\
 & \$35,000 to \$59,000 & -0.948 & -0.781\\
 & \$60,000 or more & -1.640 & -1.508\\
 & Do not know & 0.517 & 0.446\\
 & Prefer not to answer & -0.627 & -0.521\\
 & Missing & -0.310 & 0.061\\
\cmidrule{1-4}\pagebreak[0]
Household Composition & Other Family & 0.659 & 0.608\\
 & Unrelated housemates & 0.743 & 0.677\\
 & Sole Member & 0.909 & 0.833\\
 & Do not know or other & 0.069 & -0.056\\
 & Prefer not to answer & 0.443 & 0.455\\
 & Missing & 0.742 & 0.462\\
\cmidrule{1-4}\pagebreak[0]
Education Level & Secondary or equivalent & 0.011 & -0.075\\
 & Less than high school & 0.127 & 0.056\\
 & Do not know or other & 0.292 & 0.294\\
 & Prefer not to answer & 0.081 & 0.334\\
 & Missing & -0.268 & -0.156\\
\cmidrule{1-4}\pagebreak[0]
Primary Language & French & 0.114 & 0.671\\
& Other & 0.448 & 0.509\\
\cmidrule{1-4}\pagebreak[0]
LGBTQ & Non-Lgbtq & -0.031 & 0.009\\
 & Missing & 0.593 & 0.534\\
\cmidrule{1-4}\pagebreak[0]
Years in Canada & True & 0.277 & 0.102\\
\cmidrule{1-4}\pagebreak[0]
Physical Disability & True & -0.249 & -0.188\\
\cmidrule{1-4}\pagebreak[0]
Depression or Anxiety & True & 0.629 & 0.379\\
\cmidrule{1-4}\pagebreak[0]
Chronic Urinary Problem & True & 0.098 & 0.079\\
\cmidrule{1-4}\pagebreak[0]
Obesity & True & 0.145 & 0.019\\
\cmidrule{1-4}\pagebreak[0]
Personality Disorder & True & 0.179 & 0.114\\
\cmidrule{1-4}\pagebreak[0]
Stable Housing & True & 0.934 & 0.626\\
\cmidrule{1-4}\pagebreak[0]
Substance Use & True & 0.225 & 0.114\\
\cmidrule{1-4}\pagebreak[0]
Smoking or Tobacco Use & True & 0.178 & -0.063\\
\cmidrule{1-4}\pagebreak[0]
Food Insecurity & True & 0.116 & -0.106\\
\bottomrule
\label{tab:ao3-featureCoefCompareM3M4} 
\end{longtable}

\begin{longtable}[t]{lllll}
\caption{Client characteristics stratified by kernel coefficient}\\
\toprule
Variable & Values & Zero Alpha & Positive Alpha & Negative Alpha\\
\midrule
\# of Clients &  & 5038 (100\%) & 13 (100\%) & 19 (100\%)\\
\cmidrule{1-5}\pagebreak[0]
Loneliness/Social Isolation & Present & 270 (5.36\%) & 6 (46.15\%) & 0 (0.00\%)\\
\cmidrule{1-5}\pagebreak[0]
Sex & Female & 2362 (46.88\%) & 7 (53.85\%) & 10 (52.63\%)\\
 & Male & 2676 (53.12\%) & 6 (46.15\%) & 9 (47.37\%)\\
\cmidrule{1-5}\pagebreak[0]
Rural Residence & Rural & 1010 (20.05\%) & 1 (7.69\%) & 0 (0.00\%)\\
 & Urban & 3912 (77.65\%) & 11 (84.62\%) & 19 (100.00\%)\\
 & Missing & 116 (2.30\%) & 1 (7.69\%) & 0 (0.00\%)\\
\cmidrule{1-5}\pagebreak[0]
Household Income & \$0 to \$14,999 & 1240 (24.61\%) & 7 (53.85\%) & 7 (36.84\%)\\
 & \$15,000 to \$24,999 & 451 (8.95\%) & 3 (23.08\%) & 0 (0.00\%)\\
 & \$25,000 to \$34,999 & 272 (5.40\%) & 1 (7.69\%) & 1 (5.26\%)\\
 & \$35,000 to \$59,000 & 533 (10.58\%) & 1 (7.69\%) & 1 (5.26\%)\\
 & \$60,000 or more & 596 (11.83\%) & 0 (0.00\%) & 4 (21.05\%)\\
 & Do not know & 273 (5.42\%) & 0 (0.00\%) & 1 (5.26\%)\\
 & Prefer not to answer & 584 (11.59\%) & 1 (7.69\%) & 2 (10.53\%)\\
 & Missing & 1089 (21.62\%) & 0 (0.00\%) & 3 (15.79\%)\\
\cmidrule{1-5}\pagebreak[0]
Household Composition & Couple & 1886 (37.44\%) & 5 (38.46\%) & 6 (31.58\%)\\
 & OtherFamily & 516 (10.24\%) & 0 (0.00\%) & 3 (15.79\%)\\
 & Unrelated housemates & 214 (4.25\%) & 2 (15.38\%) & 1 (5.26\%)\\
 & Sole Member & 1197 (23.76\%) & 5 (38.46\%) & 3 (15.79\%)\\
 & Do not know/Other & 254 (5.04\%) & 0 (0.00\%) & 1 (5.26\%)\\
 & Prefer not to answer & 55 (1.09\%) & 1 (7.69\%) & 1 (5.26\%)\\
 & Missing & 916 (18.18\%) & 0 (0.00\%) & 4 (21.05\%)\\
\cmidrule{1-5}\pagebreak[0]
Education Level & Post-secondary or equiv & 1705 (33.84\%) & 4 (30.77\%) & 8 (42.11\%)\\
 & Secondary or equivalent & 1837 (36.46\%) & 5 (38.46\%) & 7 (36.84\%)\\
 & Less than high school & 392 (7.78\%) & 1 (7.69\%) & 2 (10.53\%)\\
 & Do not know/Other & 266 (5.28\%) & 3 (23.08\%) & 0 (0.00\%)\\
 & Prefer not to answer & 54 (1.07\%) & 0 (0.00\%) & 0 (0.00\%)\\
 & Missing & 784 (15.56\%) & 0 (0.00\%) & 2 (10.53\%)\\
\cmidrule{1-5}\pagebreak[0]
Primary Language & English & 4660 (92.50\%) & 13 (100.00\%) & 18 (94.74\%)\\
 & French & 81 (1.61\%) & 0 (0.00\%) & 1 (5.26\%)\\
 & Other & 297 (5.90\%) & 0 (0.00\%) & 0 (0.00\%)\\
\cmidrule{1-5}\pagebreak[0]
LGBTQ & Lgbtq & 66 (1.31\%) & 1 (7.69\%) & 0 (0.00\%)\\
 & Non-Lgbtq & 1077 (21.38\%) & 2 (15.38\%) & 5 (26.32\%)\\
 & Missing & 3895 (77.31\%) & 10 (76.92\%) & 14 (73.68\%)\\
\cmidrule{1-5}\pagebreak[0]
Years in Canada & True & 624 (12.39\%) & 2 (15.38\%) & 1 (5.26\%)\\
\cmidrule{1-5}\pagebreak[0]
Physical Disability & True & 239 (4.74\%) & 1 (7.69\%) & 0 (0.00\%)\\
\cmidrule{1-5}\pagebreak[0]
Depression or Anxiety & True & 408 (8.10\%) & 2 (15.38\%) & 0 (0.00\%)\\
\cmidrule{1-5}\pagebreak[0]
Chronic Urinary Problem & True & 848 (16.83\%) & 2 (15.38\%) & 2 (10.53\%)\\
\cmidrule{1-5}\pagebreak[0]
Obesity & True & 732 (14.53\%) & 1 (7.69\%) & 4 (21.05\%)\\
\cmidrule{1-5}\pagebreak[0]
Personality Disorder & True & 144 (2.86\%) & 1 (7.69\%) & 0 (0.00\%)\\
\cmidrule{1-5}\pagebreak[0]
Stable Housing & True & 549 (10.90\%) & 4 (30.77\%) & 3 (15.79\%)\\
\cmidrule{1-5}\pagebreak[0]
Substance Use & True & 745 (14.79\%) & 4 (30.77\%) & 4 (21.05\%)\\
\cmidrule{1-5}\pagebreak[0]
Smoking or Tobacco Use & True & 1443 (28.64\%) & 8 (61.54\%) & 3 (15.79\%)\\
\cmidrule{1-5}\pagebreak[0]
Food Insecurity & True & 195 (3.87\%) & 4 (30.77\%) & 1 (5.26\%)\\
\bottomrule
\label{tab:ao3-clientCharStrata}
\end{longtable}

\normalsize

\noindent Following are the top 10 codes for each topic from non-negative matrix factorization on provider type and service type (PIST2) data for sub-cohorts of clients with positive, negative, and zero \(\alpha\) coefficients.

\bigskip

\textbf{Negative \(\alpha\)}

\paragraph*{\textbf{Topic 1 with top 10 weights}}

{[}(`Diagnostic test request', 1.46), (`Intermediate assessment', 1.46), (`Physician', 1.42), (`Nurse', 1.30), (`Discussion regarding the treatment plan', 1.30), (`Health advice/instructions', 1.23), (`Case management/coordination', 1.17), (`Minor assessment', 1.03), (`Discussion regarding the diagnostic findings', 0.96), (`General assessment', 0.91){]}

\paragraph*{\textbf{Topic 2 with top 10 weights}}

{[}(`discussion', 1.41), (`Recommendation/assistance', 1.29), (`Basic support', 0.88), (`Forms completion', 0.87), (`internal referral', 0.78), (`Information provision about community resources', 0.75), (`counselling', 0.72), (`Internal consultation', 0.54), (`Case management/coordination', 0.44), (`Counselor', 0.42){]}

\paragraph*{\textbf{Topic 3 with top 10 weights}}

{[}(`Counselor', 0.75), (`Individual counselling', 0.74), (`Forms completion', 0.73), (`Foot care', 0.66), (`Chiropodist', 0.66), (`Client intake/interview', 0.59), (`Service access coordinator', 0.49), (`Blank Services (grandfathered)', 0.48), (`Preventive care', 0.47), (`medication prescription', 0.47){]}

\paragraph*{\textbf{Topic 4 with top 10 weights}}

{[}(`Periodic health examination', 1.10), (`Client intake/interview', 0.88), (`medication prescription', 0.75), (`Nurse Practitioner (RN-EC)', 0.65), (`discussion', 0.64), (`Discussion regarding the diagnostic findings', 0.46), (`Discussion regarding the treatment plan', 0.43), (`Diagnostic test request', 0.37), (`Intermediate assessment', 0.37), (`Preventive care', 0.37){]}

\paragraph*{\textbf{Topic 5 with top 10 weights}}

{[}(`Individual counselling', 1.47), (`Nurse Practitioner (RN-EC)', 1.05), (`internal referral', 0.60), (`Minor assessment', 0.58), (`External referral', 0.57), (`Dietitian/Nutritionist', 0.55), (`assessment', 0.55), (`Health advice/instructions', 0.52), (`Discussion regarding the treatment plan', 0.41), (`Medication renewal', 0.41){]}

\normalsize
\bigskip

\textbf{Positive \(\alpha\)}

\paragraph*{\textbf{Topic 1 with top 10 weights}}

{[}(`Consultation (grandfathered)', 1.25), (`Health advice/instructions', 1.12), (`referral', 1.05), (`discussion', 1.05), (`Advocacy', 1.05), (`Internal consultation', 1.05), (`Physician', 1.02), (`Nurse', 1.02), (`assessment', 0.97), (`Basic support', 0.97){]}

\paragraph*{\textbf{Topic 2 with top 10 weights}}

{[}(`External consultation', 0.74), (`External referral', 0.61), (`Minor assessment', 0.60), (`Social worker', 0.57), (`Transportation assistance', 0.49), (`Individual counselling', 0.49), (`Intermediate assessment', 0.45), (`Information provision about community resources', 0.41), (`medication prescription', 0.39), (`Community Health Worker', 0.35){]}

\paragraph*{\textbf{Topic 3 with top 10 weights}}

{[}(`Preventive care', 0.93), (`Client intake/interview', 0.68), (`Discussion regarding the treatment plan', 0.61), (`Chronic illness monitoring', 0.60), (`Discussion regarding the diagnostic findings', 0.60), (`assessment', 0.58), (`Basic support', 0.58), (`Community Health Worker', 0.57), (`care', 0.55), (`Health advice/instructions', 0.54){]}

\paragraph*{Topic 4 with top 10 weights}

{[}(`Social worker', 0.80), (`Registered Practical Nurse (RPN)', 0.80), (`Individual counselling', 0.79), (`Case management/coordination', 0.66), (`health examination', 0.59), (`Minor assessment', 0.55), (`External referral', 0.5), (`Physician', 0.46), (`Nurse', 0.46), (`Outreach Worker', 0.45){]}

\paragraph*{\textbf{Topic 5 with top 10 weights}}

{[}(`Minor assessment', 0.85), (`Outreach Worker', 0.62), (`Case management/coordination', 0.60), (`Social worker', 0.59), (`General assessment', 0.56), (`Dietitian/Nutritionist', 0.46), (`Foot care', 0.46), (`Diagnostic test request', 0.46), (`Discussion regarding the diagnostic findings', 0.46), (`Registered Practical Nurse (RPN)', 0.37){]}

\bigskip
\normalsize

\textbf{Zero \(\alpha\)}

\paragraph*{\textbf{Topic 1 with top 10 weights}}

{[}(`Health advice/instructions', 5.89), (`Nurse Practitioner (RN-EC)', 5.34), (`Discussion regarding the treatment plan', 4.96), (`Intermediate assessment', 4.85), (`Minor assessment', 4.46), (`Nurse', 4.29), (`Physician', 3.85), (`Diagnostic test request', 3.79), (`medication prescription', 3.75), (`Discussion regarding the diagnostic findings', 3.61){]}

\paragraph*{\textbf{Topic 2 with top 10 weights}}

{[}(`Basic support', 2.98), (`Advocacy', 2.90), (`Recommendation/assistance', 2.74), (`discussion', 2.58), (`counselling', 2.32), (`Consultation (grandfathered)', 2.26), (`assessment', 2.04), (`Triage', 1.85), (`referral', 1.82), (`Internal consultation', 1.63){]}

\paragraph*{\textbf{Topic 3 with top 10 weights}}

{[}(`General assessment', 2.99), (`internal referral', 2.77), (`Individual counselling', 2.56), (`Physician', 2.29), (`Internal consultation', 2.21), (`Dietitian/Nutritionist', 2.16), (`Nurse', 2.13), (`External referral', 1.93), (`Diagnostic test request', 1.81), (`Consultation (grandfathered)', 1.73){]}

\paragraph*{\textbf{Topic 4 with top 10 weights}}

{[}(`care', 2.42), (`Mental health care', 2.08), (`Preventive care', 1.93), (`Chronic illness monitoring', 1.88), (`Individual counselling', 1.85), (`Dietitian/Nutritionist', 1.40), (`assessment', 1.36), (`Blank Services (grandfathered)', 1.36), (`Dispensing medication', 1.33), (`counselling', 1.33){]}

\paragraph*{\textbf{Topic 5 with top 10 weights}}

{[}(`Information provision about community resources', 3.65), (`Community Health Worker', 2.09), (`Client intake/interview', 2.01), (`Case management/coordination', 1.99), (`Forms completion', 1.96), (`Recommendation/assistance', 1.86), (`Social worker', 1.50), (`Individual counselling', 1.24), (`internal referral', 1.23), (`Health advice/instructions', 1.21){]}


\end{document}